%% file: main.tex
\documentclass{sig-alternate}
\usepackage{times}  
\usepackage{helvet}  
\usepackage{courier}  
\usepackage{url}  
\usepackage{graphicx}  
\usepackage{amsmath}
\usepackage{physics}
\usepackage{algorithmic}
\usepackage{algorithm}
\usepackage{authblk}
\usepackage{lipsum,subcaption}

\usepackage{etoolbox}
\patchcmd{\maketitle}{\@copyrightspace}{}{}{}
 
\begin{document}
 
%
%
%
%
\title{Model Extraction Warning in MLaaS Paradigm}
\author[1]{Manish Kesarwani}
\author[2]{Bhaskar Mukhoty}
\author[1]{Vijay Arya}
\author[1]{Sameep Mehta}
\affil[1]{IBM India Research Lab}
\affil[2]{Indian Institute of Technology, Kanpur}
\renewcommand\Authands{ and }

\maketitle

\input{Abstract}
\input{Introduction}
\input{ProblemFW}
\input{ModelExtWarning}

\input{Experiments}

\input{Conclusion}

\bibliographystyle{abbrv}
\bibliography{ref}

\end{document}

%% file: Abstract.tex
\begin{abstract}
Cloud vendors are increasingly offering machine learning services as part of their platform and services portfolios. These services enable the deployment of machine learning models on the cloud that are offered on a pay-per-query basis to application developers and end users. However recent work has shown that the hosted models are susceptible to extraction attacks. Adversaries may launch queries to steal the model and compromise future query payments or privacy of the training data. In this work, we present a cloud-based extraction monitor that can quantify the extraction status of models by observing the query and response streams of both individual and colluding adversarial users. We present a novel technique that uses information gain to measure the model learning rate by users with increasing number of queries. Additionally, we present an alternate technique that maintains intelligent query summaries to measure the learning rate relative to the coverage of the input feature space in the presence of collusion. Both these approaches have low computational overhead and can easily be offered as services to model owners to warn them of possible extraction attacks from adversaries. We present performance results for these approaches for decision tree models deployed on BigML MLaaS platform, using open source datasets and different adversarial attack strategies. 
\end{abstract}

%% file: Introduction.tex
\section{Introduction}

Most cloud service providers (CSPs) now offer machine learning services that enable developers to train and host machine learning (ML) models on the cloud. 
Application developers or end users can access these models via prediction APIs on a pay-per-query basis. 
Recent work has showed that the hosted ML models are susceptible to model extraction attacks. 
Adversaries may abuse a model's query API and launch a series of intelligent queries spanning the input space to steal or replicate the hosted model, thus avoiding future query charges. Moreover, access to the model parameters can leak sensitive private information about the training data and facilitate evasion attacks especially if the model is utilized in security applications such as malware or spam classification~\cite{tramer,dwork2008differential,biggio2013evasion}.

Machine learning models generally encapsulate confidential information related to the problem domain. Additionally, the training data used to construct the model may be private and expensive to obtain. For instance, the business logic of an insurance firm is encoded in the model that predicts which customers are eligible for insurance.  Similarly, a significant number of expensive clinical trials and private health records may have been used to train a model that detects health issues in a patient. Therefore techniques to detect or prevent ongoing model extraction attacks are of significant importance to model owners and cloud service providers. Moreover, knowing the status of model extraction can help price queries as a function of extraction status and aid in making decisions regarding packaging models at an attractive price within a cloud subscription service. 

Prior work \cite{tramer} on model extraction showed that interpretable ML models such as linear and logistic regression can be replicated using a constant number of queries (of order of number of features) by solving a system of linear equations. For the non-interpretable machine learning models such as decision trees and neural networks, the number of queries needed to recover the model is variable and efficient attack strategies may be designed to successively steal the models over time. The focus of this work is to provide extraction status warnings for such machine learning models. In particular we focus on decision tree models used for classification although the proposed techniques are more broadly applicable to other machine learning models as well. We output warnings related to model extraction status for both single and multiple user query streams. In case of multiple users, our techniques can indicate the model extraction status in the presence of colluding adversaries. 
%
\begin{figure}
    \centering
\includegraphics[width=\linewidth]{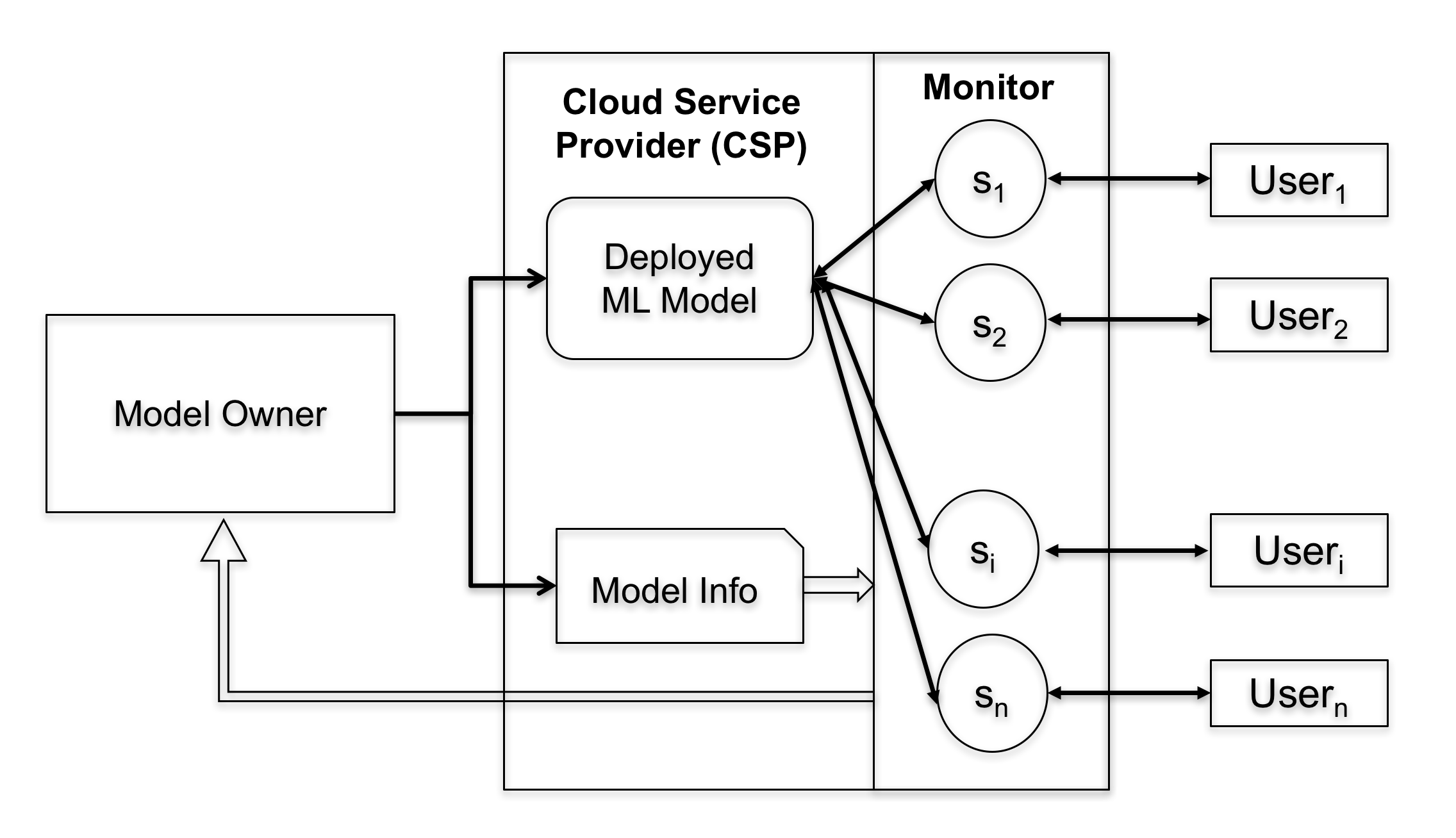}
    \caption{System Architecture}
    \label{fig_architecture}
\end{figure}

Figure~\ref{fig_architecture} depicts the problem setting with different stakeholders. Model owners deploy their models on the cloud, which are accessed by end users via pay-per-query APIs. The CSP manages the query and response streams from each user to the deployed ML models. 
In order to detect extraction attacks from adversaries, we design a cloud-based \emph{extraction monitor} that continually informs model owners of the information recoverable by users based on the queries responded thus far. The extraction monitor has at least a partial visibility of the deployed model in order to provide such a status, with estimation quality improving with better visibility. We propose two different strategies that the monitor uses to infer the model learning rate of users. In the first strategy, the monitor computes the information gain of the users based on their queries received thus far. In order to do this, it uses a validation set supplied by the model owner that has a distribution similar to the training set. In the second strategy, the monitor maintains compact summaries of user queries and estimates their coverage of the input space relative to the partition induced by the source model. Both approaches can also provide joint extraction status corresponding to queries of user groups to understand the coverage of queries from colluding users. In particular the key contributions of this work are as follows: 
\begin{itemize}
\item{
We define a new metric to measure the information learnt by a decision tree with increasing amount of training data. The metric is based on entropy and measures the learning rate of an evolving decision tree with respect to a validation set provided by the model owner. It may be used to measure the information gained by adversaries with increasing number of API queries. We present a fast greedy algorithm that identifies the maximum information gained by any set of $k$ users among all users who have queried the model, which may be utilized in the presence of colluding adversaries.  
}
\item{
We define an alternate metric that measures the information learnt by querying users by computing the coverage of the input space relative to the partitions induced by the source model owned by the owner. This approach maintains efficient summaries of user query streams and can identify the minimum number of users that have maximum information about the model in polynomial time.   
}
\item{We present experimental evaluation results for both approaches using open source datasets and decision trees deployed on BigML MLaaS platform. We present model extraction results for two attack strategies based on random queries and an intelligent path finding algorithm from prior work.}
\end{itemize}

\subsubsection{Related Work. } Most of the existing work on model extraction focuses on different types of extraction or inversion attacks that divulge model parameters or sensitive private information about the training data to adversaries querying the model~\cite{dwork2008differential,ateniese2015hacking,tramer}. 
Also model extraction can facilitate evasion in applications such as spam filtering and anomaly or malware detection~\cite{huang2011adversarial,biggio2013evasion}. 
Although work has been done to analyze adversarial frameworks, we know of no prior work that estimates the extraction status of an adversary and provides warnings to model owners, which is the focus of our work. 

The balance of the paper is organized as follows. Section~\ref{sec:pf} presents the problem framework and section~\ref{sec:mew} presents two different approaches to compute model extraction status for both individual and colluding users. The experimental results with open source datasets are presented in section~\ref{sec:exp}. Lastly, conclusions and directions for future work are presented in section~\ref{sec:conclusions}.

%% file: ProblemFW.tex
\section{Problem Framework}
\label{sec:pf}

The MLaaS paradigm allows ML developers to train a model based on their data and deploy it on the cloud. The end users query the model using its prediction APIs. We regard an ML model as a function $f : X \rightarrow y$ that maps a d-dimensional input feature vector $x_i \in X$ to a prediction $y_i \in y$, which is a categorical variable. 

\vspace{1mm}\noindent{\bf Model Extration Attack.~} In the case of a model extraction attack, an adversary abuses the prediction API of a model $f$ and attempts to learn a model $\hat{f}$ that achieves similar performance as $f$. 

\vspace{1mm}\noindent{\bf Adversary Model.~}
In our work, the adversary can be a single user or a group of colluding users that have access to the ML model's prediction API. Furthermore adversaries may have knowledge of the model type (e.g. Decision Tree or Logistic Regression), or the kind of data it is trained upon with the objective of learning the ML model's parameters via a sequence of intelligent queries. 

\vspace{1mm}\noindent{\bf Performance Metrics.~}
In order to compare the performance of a model $\hat{f}$ learnt by an adversary with respect to the deployed source model $f$, we compute the following metrics \cite{tramer}. Note that the monitor does not have access to any models learnt by the adversaries, but simply their queries. 
\subsubsection{$1-R_{test}:$}
While building the source model $f$, the model owner generally splits the dataset into train and test sets. We use this test data set $T$ to compare the performance of the source and learnt models. 
\begin{equation}
    R_{test} =\dfrac{1}{\abs{T}} \sum_{t\in T} I(\hat{f}(x^{(t)})\neq f(x^{(t)}))
\end{equation}
where $I$ denotes the indicator function. Thus $1-R_{test}$ represents the accuracy of an adversary's model with respect to the deployed model.

\subsubsection{$1-R_{unif}:$}
In addition to testing the adversary's model on the test data set that follows a specific distribution, we also test the learnt model against a data set $U$ that is generated uniformly from the input feature space. 
\begin{equation}
    R_{unif} =\dfrac{1}{\abs{U}} \sum_{t\in U} I(\hat{f}(x^{(t)})\neq f(x^{(t)}))
\end{equation}
Thus $1-R_{unif}$ represents another metric to evaluate the performance an adversary's model as compared to a deployed model. 

%

%% file: ModelExtWarning.tex
\section{Model Extraction Warning}
\label{sec:mew}

Our goal is to design a cloud based monitor capable of providing model extraction status and warnings. 
Each query to a deployed ML model $f$ along with its response leaks a certain amount of information about the decision boundaries of the model to the user. 
After having made multiple queries, an adversarial user can train his own ML model $\hat{f}$ over the query-response set, which has a certain performance accuracy as compared to the source model $f$. 
Our objective is to detect if any set of clients of size $k \ge 1$ can jointly reconstruct a model that yields an accuracy beyond a given threshold. 
The model owner may run the detection at fixed time intervals or after a certain number of queries have been answered by the deployed model.
In the subsequent sections, we focus on the decision tree models for classification and propose two strategies to detect model extraction by adversarial users. 

\subsection{Strategy 1: Model Extraction Warning using Information Gain. }
\label{subsec:s1}

\subsubsection{Initialization.~}
\label{subsubsec:s1init}
In this strategy, we assume that the model owner has provided an extraction threshold $t$ and a validation set to the monitor. The monitor observes the query response pairs of each user and incremently learns a local  decision tree based on these pairs. 
Let $S$ denote the validation set, with each element having an input $x$ and an associated class label $y\in[k]$. Let $p_i$ denote the probability that a random element chosen from $S$ belongs to the class $i$. 
Then the entropy of $S$ is defined by: 
\begin{equation}
      Entropy(S)=-\sum\limits_{i=1}^{k}p_i \log{p_i}
\end{equation}

\subsubsection{Information gain of a decision tree.~} 
\label{subsubsec:s1gain}
Recall that the information gain of an attribute is defined as the reduction in entropy of a training set when we partition it based on values that the attribute can take~\cite{quinlan1986induction}. 
%
%
%
%
We generalize this concept and define a metric called the \emph{information gain of a decision tree} $T$ with respect to a given validation set $S$. 
This metric is essentially computed by the monitor on its local decision tree to estimate the information gained by users based on their queries thus far. 
By evaluating $S$ on $T$, let $S$ be partitioned into disjoint subsets $\{S_l\}$ by the leaf nodes of $T$. 
The information gain of $T$ is defined as the reduction in entropy as follows: 
  \begin{equation}
      IG_{tree}(S,T)=Entropy(S)-\sum\limits_{l\in leaf(T)}\frac{\abs{S_l}}{\abs{S}}Entropy(S_l)
      \label{eq:info_gain}
  \end{equation}
%
where the entropy of a leaf node $l$ i.e. $Entropy(S_l)$ is computed using the original class labels of the validation elements.
Similarly, one can apply Eq.~\eqref{eq:info_gain} to compute the information gain of the model owners source decision tree $T_O$ as $IG_{tree}(S,  T_O)$. 

\begin{figure}
    \centering
\includegraphics[scale=0.19]{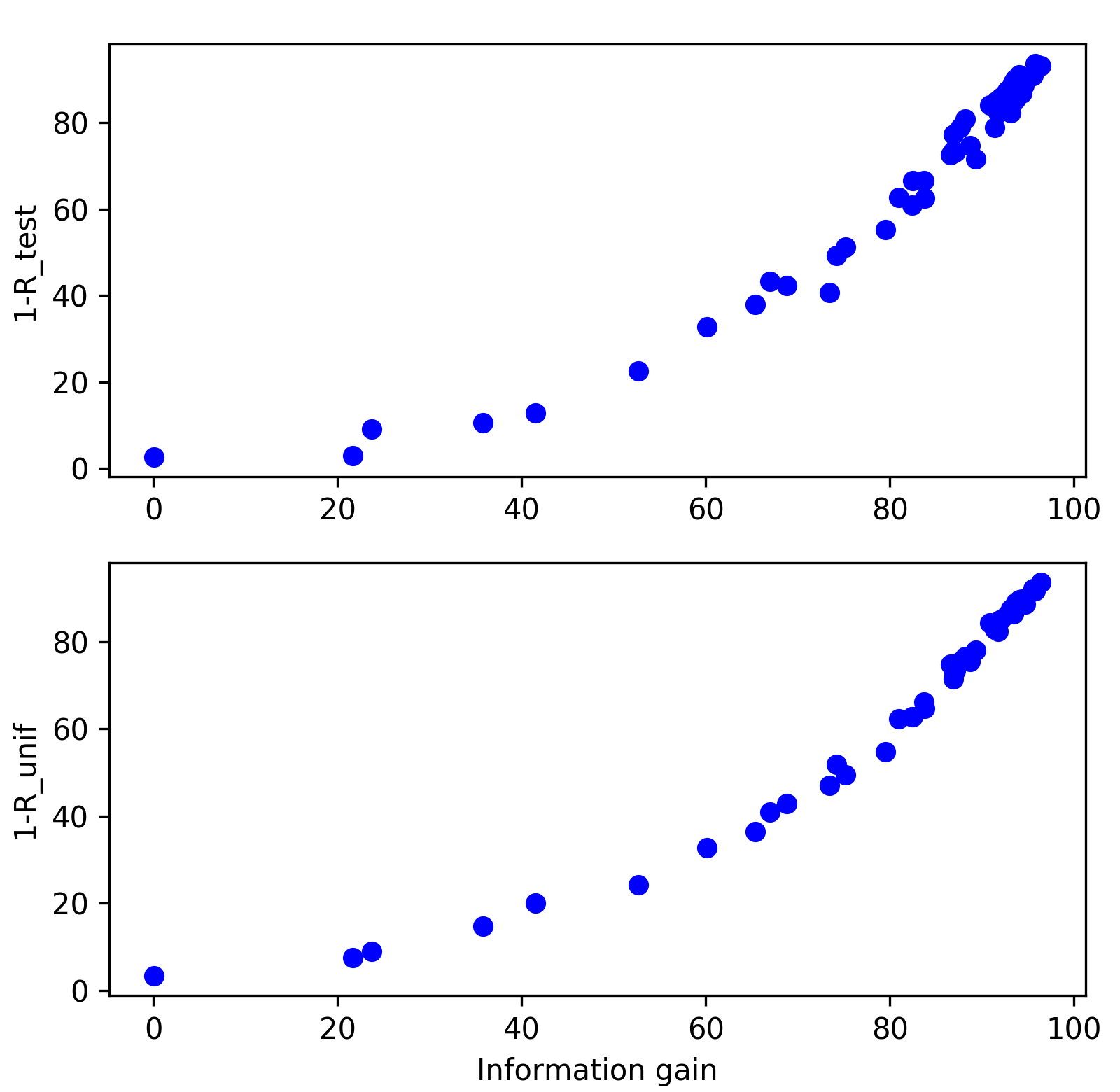}
    \caption{Accuracy vs Information Gain}
    \label{fig:acc_vs_info_gain}
\end{figure}

Figure \ref{fig:acc_vs_info_gain} plots the accuracy of a user's decision a tree with respect to its information gain with increasing amount of training data. We observe that information gain accurately captures the learning rate of the user's model. 

%

%
%

\subsubsection{Model Extraction Status Per User.~}
\label{subsubsec:userstat} 
To compute the model extraction status for each user, the monitor trains a local decision tree $T_{u}$ based on all query-response pairs made by user $u$ to the source decision tree $T_O$. 
It uses the validation set $S$ to compute the information gain $IG_{tree}(S, T_{u})$ using eq.\eqref{eq:info_gain} and the extraction status as follows: 
\begin{equation}
    Extraction\_Status_{u} = \frac{IG_{tree}(S,T_{u})}{IG_{tree}(S,T_O)} * 100
\end{equation}

For all the users whose $Extraction\_Status$ is more than the threshold $t$, a model extraction warning is generated and sent to the relevant model owners. 

\subsubsection{Model Extraction Status with Collusion.}
\label{subsubsec:usercoll}
In a general setting, a model owner would like to know if any $k$ of the $n$ users accessing the deployed model are colluding to extract it. We extend our existing approach to compute the model extraction warning in case of collusion as follows.

In a brute force approach, a monitor would need to compute the extraction status for all $n \choose k$ combinations of users. 
Algorithm \ref{alg:comb} shows how the monitor can recover a set of $k$ users that have maximum knowledge of the deployed model using their combined query data. In this algorithm the function call $DecisionTree(query\_points)$ returns a decision tree trained over the provided $query\_points$.
\begin{algorithm}
  	\caption{$combUserSelection$}
  	\label{alg:comb}
  	\begin{algorithmic}
  		\renewcommand{\algorithmicrequire}{\textbf{Input:}}
  		\renewcommand{\algorithmicensure}{\textbf{Output:}}
  		\REQUIRE validation set $S$, query history $Q$, parameter $k$
  		\ENSURE set of $k$ users, their extraction status
  		\STATE $source\_gain = IG_{tree}(S,T_O)$
  		\STATE $pool=combinations(n,k)$
  		\FOR{$c \in pool$}
  		    \STATE $query\_points = \{\}$
  		    \FOR{$user \in c$}
  		        \STATE $query\_points=\{query\_points\cup Q[user]\}$
  		    \ENDFOR
  		
  		\STATE $gain[c]=IG_{tree}(S,DecisionTree(query\_points))$
  		\ENDFOR
        \STATE $U=argmax(gain)$
        \STATE $max\_gain=\max(gain)$
        \RETURN $U, (max\_gain/source\_gain)*100$
  	\end{algorithmic}
  \end{algorithm}
  
The complexity of this algorithm is exponential in $n$.
To overcome the issue of exponential running time, we propose a greedy algorithm, where at each step the monitor selects that user whose query set in combination with the queries of selected users yields a decision tree that has the highest information gain. Algorithm \ref{alg:greedy} gives the pseudo-code for the greedy method. The algorithm needs a parameter $k$ and the query history of all $n$ users as input and returns a set of $k$ users that have the highest information along with their extraction status. 
\begin{algorithm}
  	\caption{$greedyUserSelection$}
  	\label{alg:greedy}
  	\begin{algorithmic}
  		\renewcommand{\algorithmicrequire}{\textbf{Input:}}
  		\renewcommand{\algorithmicensure}{\textbf{Output:}}
  		\REQUIRE validation set $S$, query history Q, parameter $k$
  		\ENSURE set of $k$ users, their extraction status
  		\STATE $source\_gain = IG_{tree}(S,T_O)$
  		\STATE $pool=[1, \cdots ,n], U=\{\}$
  		\FOR{$i = 1 to k$}
  		    \STATE $x=-1, max\_gain=0$
  		    \FOR{$j \in pool$}
  		        \STATE $gain = IG_{tree}(S,DecisionTree(Q[\{U \cup j\}]))$
  		        \IF{$gain > max\_gain$}
  		        \STATE $x=j, max\_gain=gain$
  		        \ENDIF
  		    \ENDFOR
  		\IF{$x!=-1$}
  		    \STATE $U=\{U\cup x\}, pool=\{pool \setminus x\} $
  		\ENDIF
  		\ENDFOR
  		
        \RETURN $U, (max\_gain/source\_gain) * 100$
  	\end{algorithmic}
  \end{algorithm}

\subsubsection{Quality of Model Extraction Warning. }
\label{subsubsec:s1quality}
We present experimental results of model extraction warnings on real datasets in Section \ref{sec:exp}, which shed light on the accuracy of the extraction status. One of the factors that affects the quality of warning is the classification accuracy of the validation set. 
If the validation set is not accurately classified  by $T_O$ then, it may contains less information and all the samples in the validation set may belong to a few classes only. To avoid this, the validation set may be assigned an information content score as follows:
\begin{equation}
    Score(S)=IG_{tree}(Tr,DecisionTree(S))/Entropy(Tr)
\end{equation}
where $Tr$ is the training set confidential to the model owner. The model owner scores the validation set using the above rule before handing it to the monitor. The score can thus be used to assign a confidence value to the model extraction status estimates. 


\begin{algorithm}
  	\caption{$updateModelSummaries$}
  	\label{alg:userSummary}
  	\begin{algorithmic}
  		\renewcommand{\algorithmicrequire}{\textbf{Input:}}
  		\renewcommand{\algorithmicensure}{\textbf{Output:}}
  		\REQUIRE $UserID$, Query $q$
  		\ENSURE create/update user summary
  		\STATE $S = getExistingUserSummary(UserID)$
  		\STATE $classID, leafID \leftarrow T_{S}.predict(q)$
  		\IF {$S~ != Null $}
  			\STATE $leaf\_info \leftarrow S[leafID]$
  			\FOR {$feature  \in leaf\_info$}
  				\STATE $feature[0] = min(q[feature],feature.min)$
  				\STATE $feature[1] = max(q[feature],feature.max)$
  			\ENDFOR	
  		\ELSE
  			\STATE $leaf\_info[f_t][2] \leftarrow newArray()$
  			\FOR {$feature  \in leaf\_info$}
  				\STATE $feature[0] = q[feature]$
  				\STATE $feature[1] = q[feature]$
  			\ENDFOR
  		\ENDIF  		
  		\STATE $S[leafID] \leftarrow leaf\_info$
  	\end{algorithmic}
  \end{algorithm}
  
\subsection{Strategy 2: Model Extraction Warning using Compact User Summaries }
\label{subsec:s2}


\begin{figure*}[!t]
  \centering
  \subcaptionbox{Class Boundaries of Source Model}[.23\linewidth][c]{%
    \includegraphics[width=.2\linewidth]{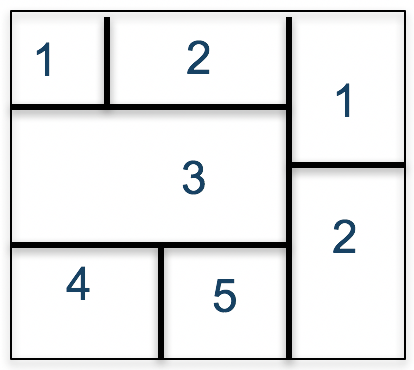}}\quad
  \subcaptionbox{Summary of User A}[.23\linewidth][c]{%
    \includegraphics[width=.2\linewidth]{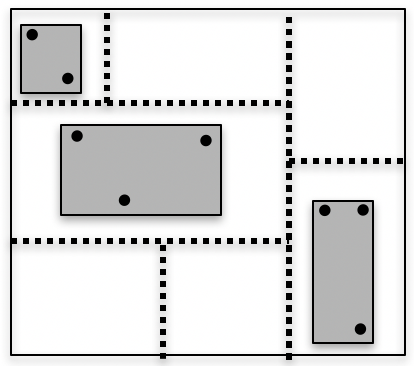}}\quad
  \subcaptionbox{Summary of User B}[.23\linewidth][c]{%
    \includegraphics[width=.2\linewidth]{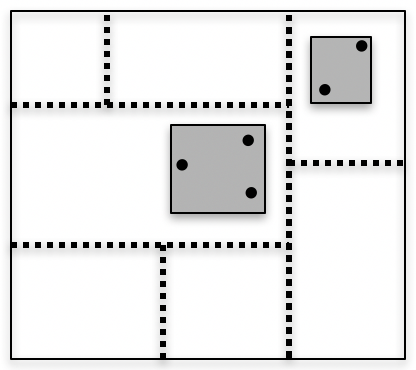}}\quad
  \subcaptionbox{Combined Summary of User A and B}[.23\linewidth][c]{%
    \includegraphics[width=.2\linewidth]{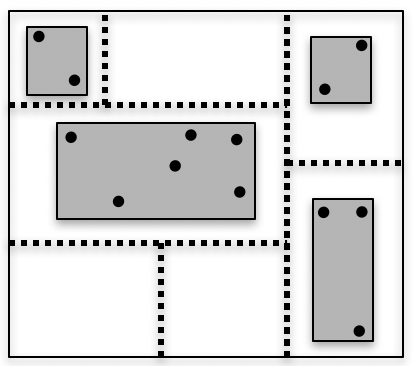}} 
  \caption{Model Summary Example}
  \label{fig:modelsummary}
\end{figure*}
\subsubsection{Initialization}
\label{subsubsec:s2}
In the previous strategy, the monitor stores historical queries of all users accessing a model and maintains running decision trees on a combination of these queries. This can be computationally expensive if the number of users and queries are large. We now propose a novel approach wherein the monitor maintains and updates a compact model summary corresponding to each user with increasing number of queries. Each user's model summary effectively encompasses boundaries of regions within the feature space that the user may have learnt for each class. The monitor assess the coverage of summaries within their respective classes to compute the overall learning rate of a user. To assess the coverage accurately, the monitor uses the leaf id to which a query maps in the source decision tree in addition to the class label, which most decision tree APIs include in their response \footnote{This assumption is not new as authors in \cite{tramer} leverage it to mount model extraction attacks against decision trees.} (e.g. BigML). Additionally,  if there are $C$ classes, the monitor obtains the class probabilities from the model owner i.e. probability $p_i$ that an element from the training data belongs to class $C_{i}$ and the hyper volume of each class ($Class\_Vol_{i}$) (explained in the next subsections). 


\subsubsection{Model Summary for each user. }
\label{subsubsec:s2extuser}
%
For each user, the monitor observes all queries and their responses made to the deployed model $T_O$. The query response includes the class label and the leaf id to which a query maps in $T_O$. For a deployed decision tree model $T_O$ with $l_{t}$ leaf nodes and $f_t$ continuous input features\footnote{This strategy does not consider categorical features, since the notion of area or hyper-volume does not hold for such features}, each user's model summary is maintained as a matrix $M$ of size $l_t \times f_t $. The entry $M(i, j)$ corresponds to a linear constraint ($[min,max]$) on the feature $j$ that the user has learned for leaf node $i$. When a user issues a new query to $T_O$, its summary is updated using Algorithm \ref{alg:userSummary}. 

For ease of exposition, we present a simple example. Consider a deployed decision tree model with $f_t=2$ training features, $t=7$ leaf nodes, and $5$ classes. Let the class labels and decision boundaries created by the leaf nodes be as shown in Figure~\ref{fig:modelsummary}(a). Let two users $A$ and $B$ query this deployed model. Then, Figure \ref{fig:modelsummary}(b) and \ref{fig:modelsummary}(c) show the user queries as points in the feature space along with the rectangular summaries stored by the monitor for each leaf node. 

  
 \subsubsection{Model Extraction Status Per User. }
\label{subsubsec:s2extstat}
%
 

To compute model extraction status for a user, the monitor computes the area (equivalently hypervolume when $f_t >= 3$) corresponding to each leaf $i$ using entries $M(i, j)~\forall j$ in the model summary matrix as follows,

\begin{equation}
Leaf\_Vol_{i} = \prod_{j \in f_t } abs(M(i,j).max - M(i,j).min) ~~\forall i \in l_t
\end{equation}

Then the area of each class is computed by aggregating leaves of the same class and the extraction percentage per class is computed using the ratio of the area in the model summary to the area per class of the source model (Note that the monitor has access to the hypervolume of each class in the source model) as shown in Equation \ref{eq:s2ExtPerClass}.

\begin{equation}
\label{eq:s2ExtPerClass}
Ext\_Per\_Class_i = \frac{\sum_{l \in C_i}(Leaf\_Vol_{i})}{Class\_Vol_i} ~~ \forall i \in C
\end{equation}

Finally, the extraction status of a user is computed as a weighted sum of per-class extraction percentages by weighing each term with the class probability $p_i$. 

\begin{equation}
\label{eq:s2extstatus}
Extraction\_Status = \sum_{i \in C}(p_i * Ext\_Per\_Class_i)
\end{equation}

For all the users whose extraction status is above the threshold $t$, a model extraction warning is generated and sent to the relevant model owners.

\subsubsection{Model Extraction Status with Collusion. }
\label{subsubsec:s2coll}
As discussed for Strategy 1 in Section \ref{subsubsec:usercoll}, a model owner may be interested to know the joint extraction of a set of $k$ users who may be colluding. Here again we use a greedy approach to compute the set of $k$ users with maximum knowledge of the model in polynomial time. The only difference in the algorithm is that during the the user selection step, the monitor chooses a user such that the combined model summary has the maximum hypervolume instead of maximum Information Gain. Fig.~\ref{fig:modelsummary}(d) illustrates an example of combining the model summaries of two users. The next section presents experimental results of extraction status for both single and collusion of adversarial users.

%% file: Experiments.tex
\section{Experiments} 
\label{sec:exp}
\begin{figure*}[h!]
  \centering
  \subcaptionbox{IRS Tax}[.24\linewidth][c]{%
    \includegraphics[width=.24\linewidth, height=7cm]{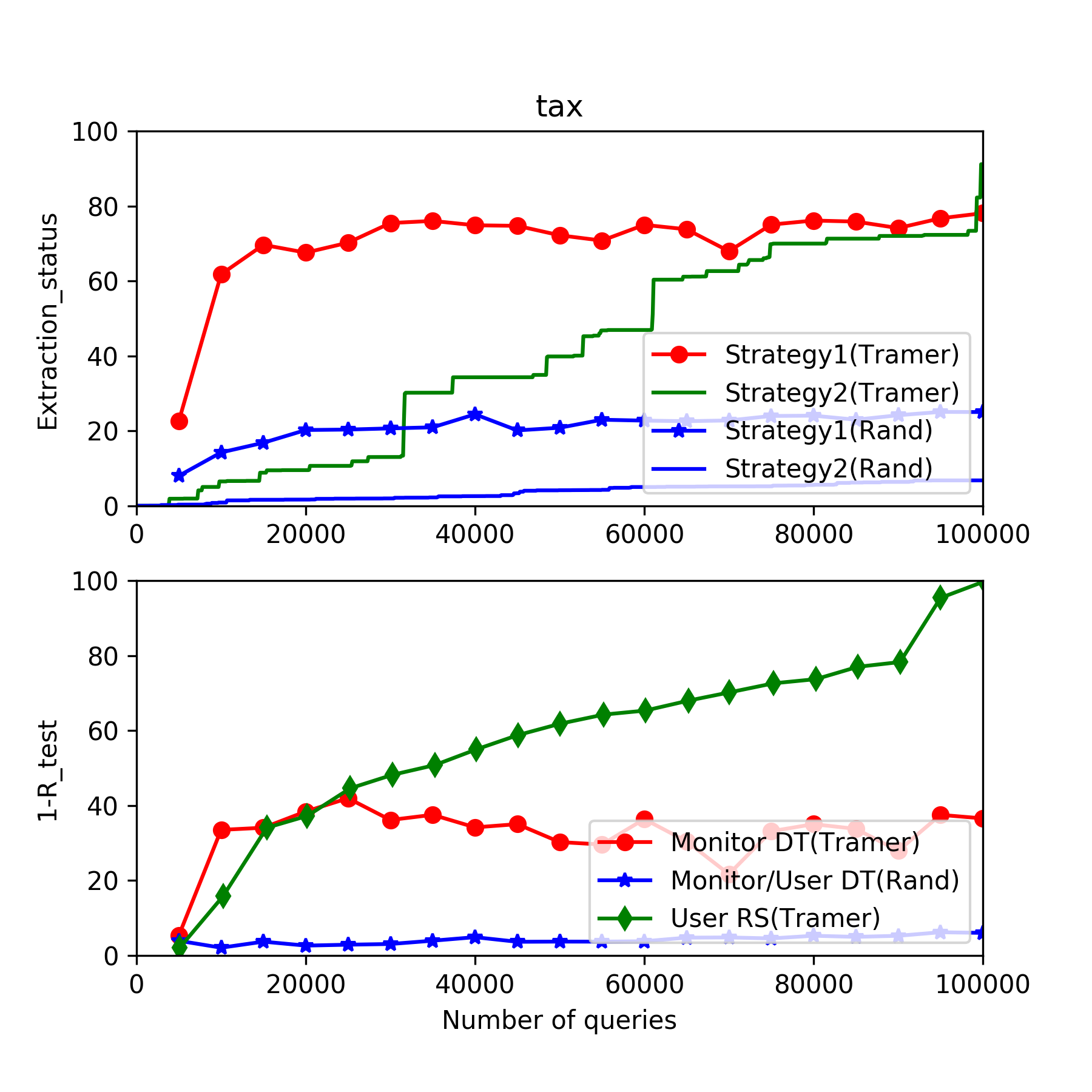}}
  \subcaptionbox{GSS Survey}[.24\linewidth][c]{%
    \includegraphics[width=.24\linewidth, height=7cm]{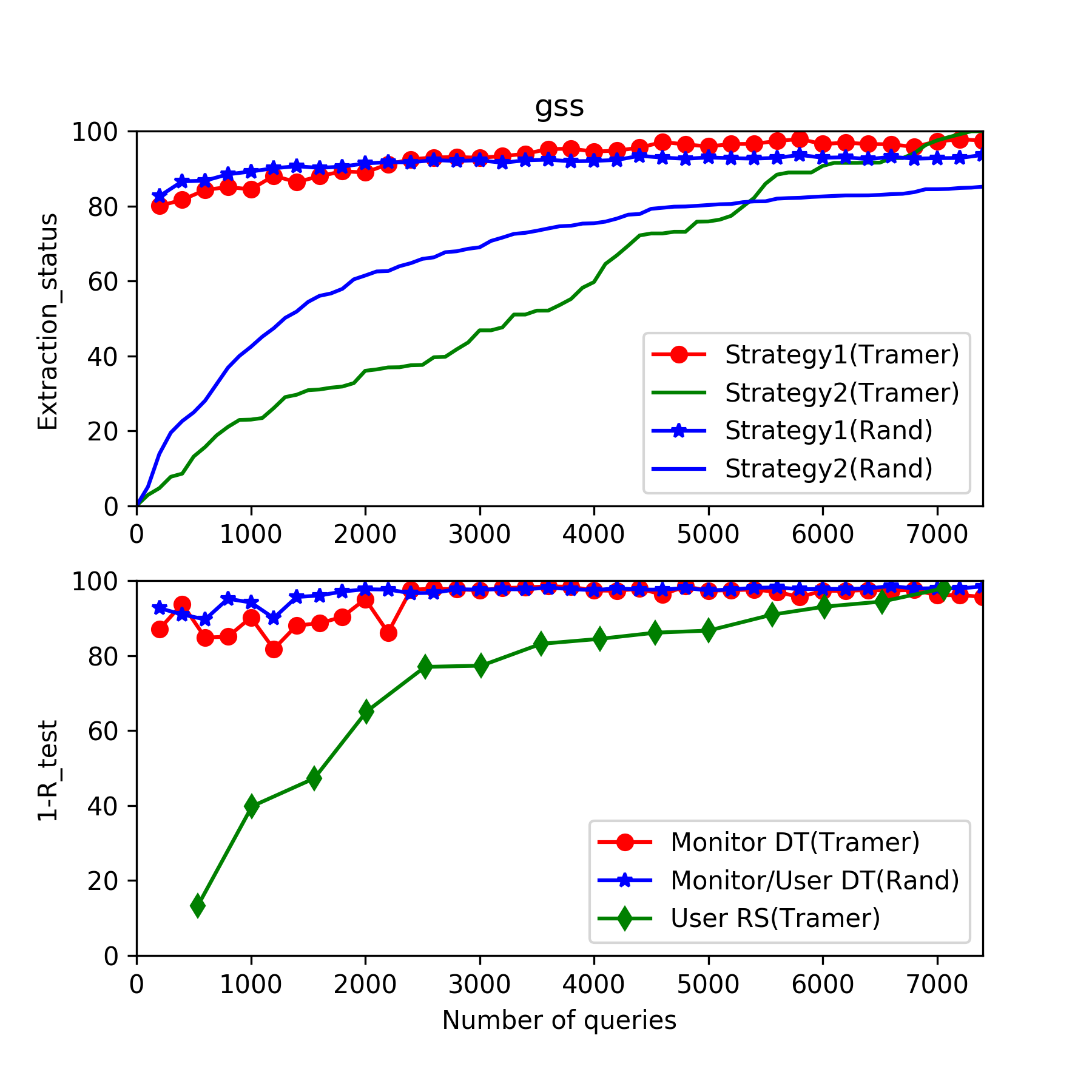}}
  \subcaptionbox{Email Importance}[.24\linewidth][c]{%
    \includegraphics[width=.24\linewidth, height=7cm]{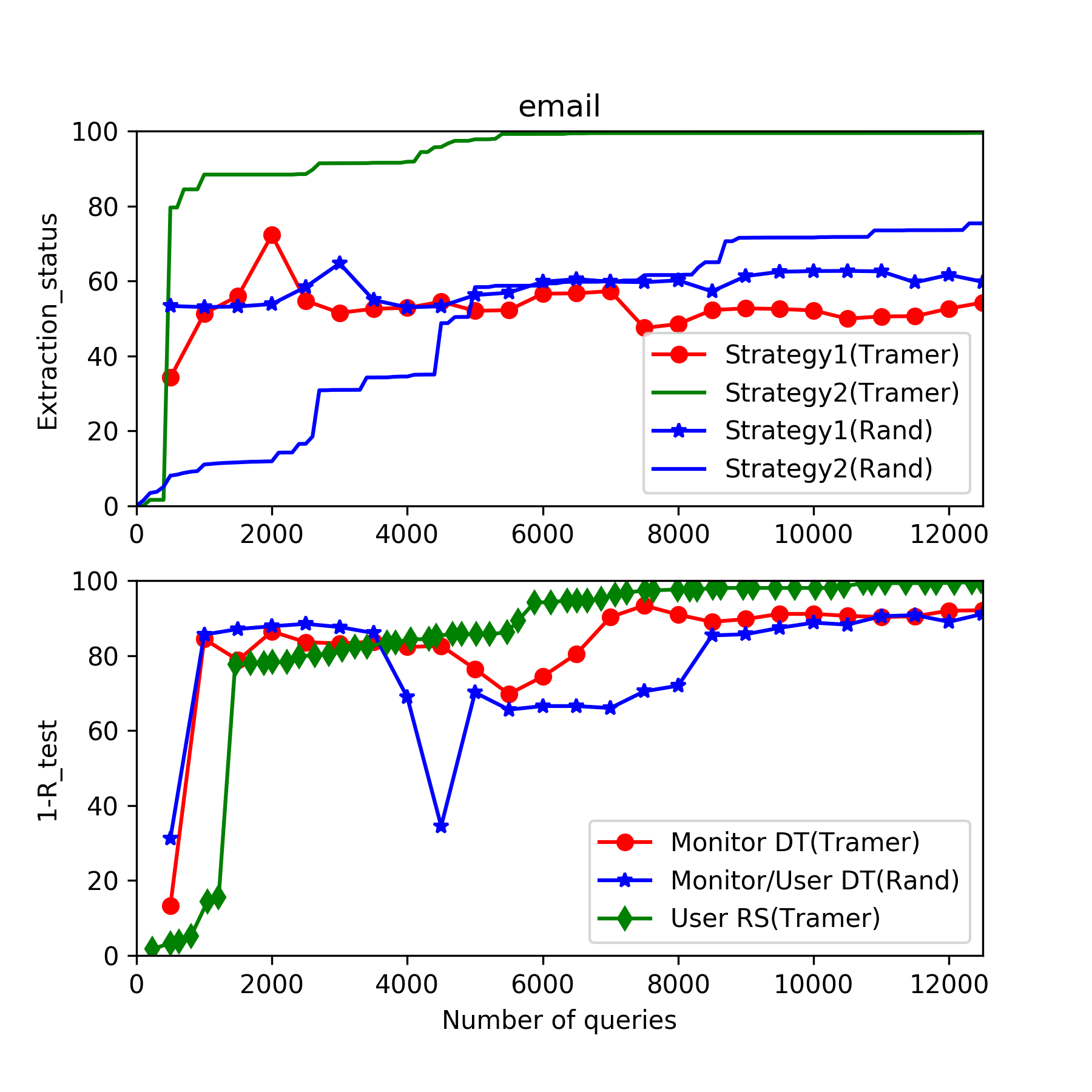}}
  \subcaptionbox{Steak Survey}[.24\linewidth][c]{%
    \includegraphics[width=.24\linewidth, height=7cm]{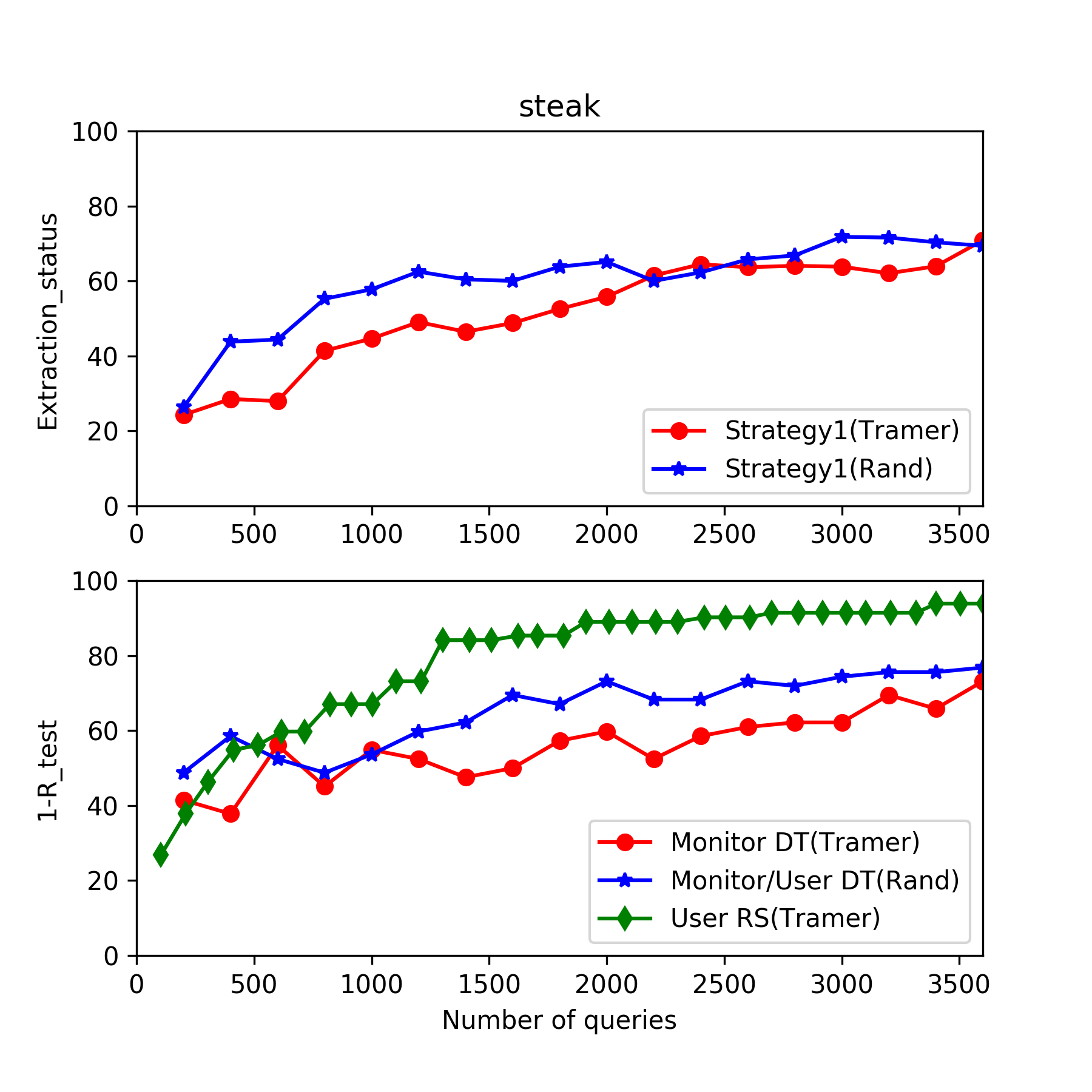}}
  \caption{Results of model extraction status for individual adversaries based on BigML datasets.}
  \label{fig:modelsummary}
\end{figure*}

%
%
%
%
%

In this section we present experimental results to evaluate the performance of the two model extraction warning strategies outlined in section \ref{sec:mew}. The results are presented for both individual as well as colluding adversaries. We simulate two attack strategies that are employed by adversaries: (i) \emph{Random query attack}: This is a naive strategy wherein adversaries generate random queries spanning the input space and get their prediction from the deployed model. Further these adversaries use this query-response pair to train a decision tree in order to steal the deployed model. (ii) \emph{Decision tree path-finding algorithm (Tramer attack)}: This is a more sophisticated attack approach proposed by~\cite{tramer} to specifically steal decision tree models. Here an adversary generates input samples to recursively search for the rule sets corresponding to each leaf node of the deployed tree. It leverages the leaf ids which are returned by the query response in addition to the class label.  To remain consistent with prior work and present extraction status results for ML models deployed as MLaaS, we use BigML's decision tree APIs. The model owner uses these API to train and deploy a decision tree. The adversaries abuses this API to launch extraction attacks.  We assume that adversaries are aware of the feature space in order to generate intelligent queries. However, the monitor is agnostic to any attack strategies employed by the adversaries. 

We use the BigML datasets listed in table~\ref{tab:datasets} for experimentation as used in prior work. The IRS model predicts a US state, based on administrative tax records. The Steak and GSS models respectively predict a person's preferred steak preparation and happiness level, from survey and demographic data. The Email Importance model predicts whether Gmail classifies an email as 'important' or not, given message metadata~\cite{tramer}.  The original decision tree models have been trained and deployed on BigML by authors of~\cite{tramer}. The validation set for strategy 1 is chosen randomly from training data and varies between $5$-$10$\% of the dataset size. 

%

\begin{table}
\begin{center}
 \begin{tabular}{| c | c | c | c |} 
 \hline
 Dataset & Records & Features & Classes \\ 
 \hline
 IRS Tax Pattern & 191283 & 37 & 51 \\ 
 \hline
 GSS Survey & 51020 & 7 & 3 \\
 \hline
 Email Importance & 4709 & 14 & 2 \\
 \hline
  Steak Survey & 412 & 12 & 5 \\
 \hline
\end{tabular}
\end{center}
\caption{BigML Datasets used for experimentation}
\label{tab:datasets}
\end{table}

%
%


%
%
%
%
%

\begin{figure*}[h!]
  \centering
  \subcaptionbox{GSS Survey}[.48\linewidth][c]{%
    \includegraphics[width=.48\linewidth, height=7cm]{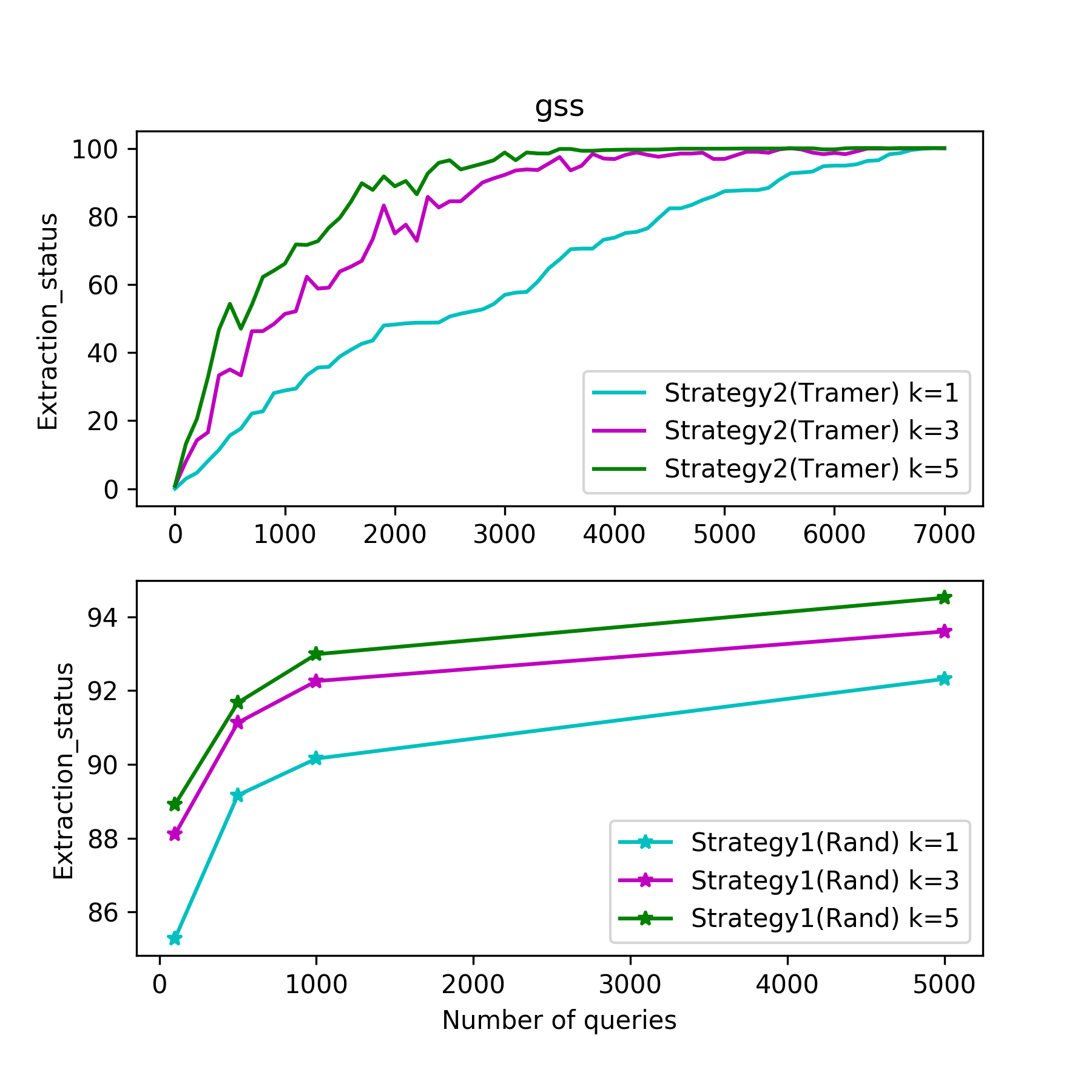}}\hspace{1mm}
  \subcaptionbox{Email Importance}[.48\linewidth][c]{%
    \includegraphics[width=.48\linewidth, height=7cm]{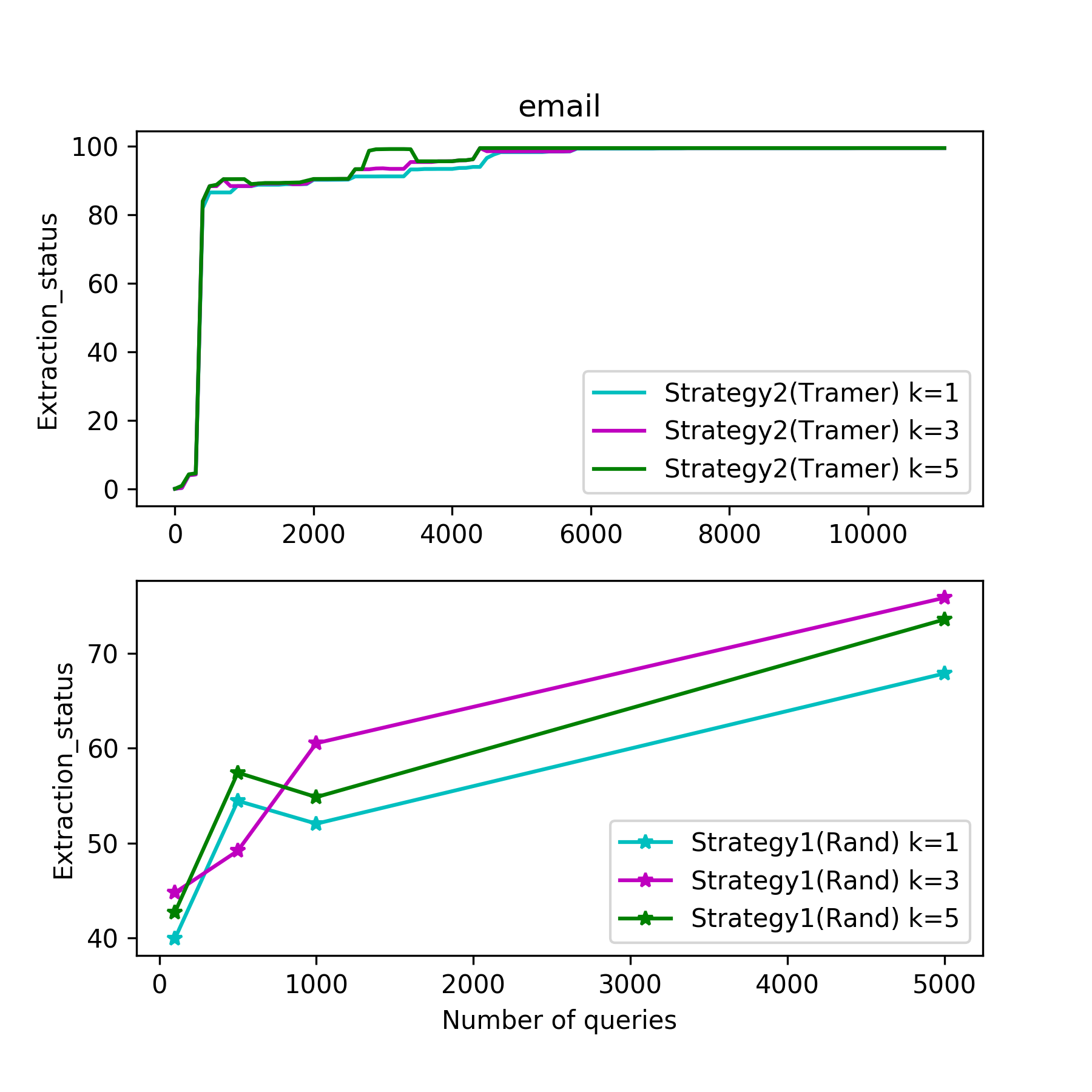}}\hspace{1mm}    
  \caption{Results of model extraction status for colluding adversaries based on BigML datasets.}
  \label{fig:collusion}
\end{figure*}

\subsubsection{Model extraction status for individual adversaries.}
\label{subsubsec:indvidual}
Figure~\ref{fig:modelsummary} plots the extraction status output by the monitor using the two proposed strategies and compares these with the performance of the model learnt by the adversary using $1-R_{test}$ with increasing number of queries. In each case, we observe that the $1-R_{test}$ steadily increases when the adversary launches the Tramer attack to recover the rule sets of all leaf nodes (i.e. ``User RS(Tramer)"). However the performance of an adversary's decision tree learnt from random queries varies across datasets as the distribution of random queries and training data may vary (i.e. ``Monitor/User DT(Rand)"). 
In strategy 1, the monitor maintains a running decision tree to compute the information gain with respect to the validation set. 
When strategy 1 is used to estimate learning rate of the user model based on Tramer attack, we observe that the initial extraction status is high (``Strategy 1(Tramer)") as the Tramer attack only recovers the rule sets of leaves sequentially without attempting to utilize any other benefit from the generated queries. 
In contrast, the monitor attempts to quantify the knowledge gained from the generated queries using a decision tree and therefore estimates a higher extraction status. The $1-R_{test}$ performance of monitor's decision tree is shown under ``Monitor DT(Tramer)", while the corresponding extraction status is shown under ``Strategy1(Tramer)". 

For higher number of queries, monitor's extraction status based on strategy 1 better matches the $1-R_{test}$ of the adversary's model based on Tramer attack (i.e. ``User RS(Tramer)"). We also observe that the $1-R_{test}$ of the decision tree constructed by the monitor can sometimes yield a lower value due to the difference in distribution of training set and the queries generated by Tramer attack. 

In case of strategy 2, we observe that the extraction status more closely matches the $1-R_{test}$ for models learnt by the adversary using random queries (``Strategy 2 (Tramer)") as well as Tramer attack queries (``Strategy 2 (Tramer)" ). In this case, the monitor maintains compact model summaries based on the span of the input space and therefore is less impacted by the distribution of queries generated by either Tramer attack or Random queries. Figure~\ref{fig:modelsummary}(d) plots results only for strategy 1 as the dataset includes several categorical features while strategy 2 is more suited for continuous features. 


\subsubsection{Model extraction status for colluding adversaries.}
\label{subsubsec:collusion}

Figure~\ref{fig:collusion} plots the extraction status output by the monitor for the case of colluding adversarial users. 
The top subplots present results for the setting where $n = 10$ users launch simultaneous Tramer attacks. The model extraction status is generated using strategy 2 when any $k \in \{1,3,5\}$ users are colluding ($k=1$ implies single adversarial user). As expected, when the number of colluding users increase, the extraction status increases consistently during all times of the attack. For example, Figure \ref{fig:collusion}(a) shows that when $1000$ queries are processed from each user, the extraction status for single adversarial user ($k = 1$) is $28\%$, for $k = 3$ users is $51\%$, and for $k=5$ users is $69\%$. 

Similarly the bottom plots consider the setting of $n = 10$ users launching simultaneous Random attacks. The model extraction status is generated using strategy 1 when any $k \in \{1,3,5\}$ users are colluding. As discussed above, a similar increase in the extraction status occurs as the number of colluding users increases. 

%

%% file: Conclusion.tex
\section{Conclusion and Future Work}
\label{sec:conclusions}

More and more cloud vendors such as Amazon, Google, and IBM are offering next generation services based on machine learning. While these services are gaining popularity among application developers, they are susceptible to attacks from adversaries who may steal the deployed models and compromise future payments or privacy of the training data. 

In this work, we presented the design of a cloud-based extraction monitor that can inform model owners about the status of model extraction by both individual and colluding adversaries in the context of decision trees. We proposed two novel metrics to infer the model learning rate of adversarial users. The first metric is based on entropy and measures the information gain of a decision tree with respect to a validation set provided by the model owner. The second metric is based on maintaining compact model summaries and computing the coverage of the input space relative to the partitions induced by the source model. Both these metrics may also be used within a greedy algorithm to determine the set $k$ users who have maximum knowledge of the model in the presence of collusion. We evaluated these metrics using two known attack techniques on BigML datasets. Our experimental results show that these metrics can provide approximate information about the knowledge extracted by adversarial users in the context of decision tree models. 

In future work, we plan to extend our work to more datasets and other non-interpretable ML models such as neural networks. In particular, we plan to explore the use of proxy ML models at the monitor that can provide extraction status of original source models irrespective of their type. We also plan to explore information theoretic metrics to quantify the loss of privacy of training data in model extraction attacks. 